%% file: PRLetters_SigNet.tex
\documentclass[times,twocolumn,final]{elsarticle}

\usepackage{prletters}
\usepackage{framed,multirow}

\usepackage{amsmath,amssymb}
\usepackage{latexsym}

\usepackage{url}
\usepackage{xcolor}
\definecolor{newcolor}{rgb}{.8,.349,.1}

\journal{Pattern Recognition Letters}

\newcommand{\etal}{\textit{et al.}}
\newcommand{\ie}{\textit{i.e.}}
\newcommand{\viz}{\textit{viz.}}

\newcommand{\eq}[1]{Eqn.~\ref{#1}}
\newcommand{\fig}[1]{Figure~\ref{#1}}
\newcommand{\sect}[1]{Section~\ref{#1}}
\newcommand{\tab}[1]{Table~\ref{#1}}

\newcommand{\nCr}[2]{\,^{#1}C_{#2}} % nCr
\DeclareGraphicsExtensions{.pdf,.png}

\begin{document}

% extra texts in front 
%\input{./tex/extras.tex}

\begin{frontmatter}

\title{SigNet: Convolutional Siamese Network for Writer Independent Offline Signature Verification}

\author[1]{Sounak \snm{Dey}\corref{cor1}} 
\cortext[cor1]{Corresponding author: 
  Tel.: +34 93 581 18 28;  
  Fax: +34 93 581 16 70;}
\ead{sdey@cvc.uab.es}
\author[1]{Anjan \snm{Dutta}}
\author[1]{J. Ignacio \snm{Toledo}}
\author[1]{Suman K.\snm{Ghosh}}
\author[1]{Josep \snm{Llad\'{o}s}}
\author[2]{Umapada \snm{Pal}}

\address[1]{Computer Vision Center, Computer Science Dept., 
Universitat Aut\`{o}noma de Barcelona, 
Edifici O, Campus UAB, 08193 Bellaterra, Spain}
\address[2]{Computer Vision and Pattern Recognition Unit,
Indian Statistical Institute, 
203, B. T. Road, Kolkata-700108, India}

% \received{1 May 2013}
% \finalform{10 May 2013}
% \accepted{13 May 2013}
% \availableonline{15 May 2013}
% \communicated{S. Sarkar}

% Abstract
\input{abstract.tex}

% Keyword

\end{frontmatter}

%\linenumbers
% \begin{verbatim} 
%   \usepackage{prletter}
% \end{verbatim}
%% main text

\input{introduction.tex}
\input{signet.tex}

% Experiments
\input{experiments.tex}

% Conclusions
\input{conclusions.tex}

% Acknowledgement
\input{acknowledgement.tex}

\bibliographystyle{elsarticle-num}
\bibliography{bibliography}

\end{document}

%% file: abstract.tex
\begin{abstract}
Offline signature verification is one of the most challenging tasks in biometrics and document forensics. Unlike other verification problems, it needs to model minute but critical details between genuine and forged signatures, because a skilled falsification might only differ from a real signature by some specific kinds of deformation. This verification task is even harder in writer independent scenarios which is undeniably fiscal for realistic cases. In this paper, we model an offline writer independent signature verification task with a convolutional Siamese network. Siamese networks are twin networks with shared weights, which can be trained to learn a feature space where similar observations are placed in proximity. This is achieved by exposing the network to a pair of similar and dissimilar observations and minimizing the Euclidean distance between similar pairs while simultaneously maximizing it between dissimilar pairs. Experiments conducted on cross-domain datasets emphasize the capability of our network to handle forgery in different languages (scripts) and handwriting styles. Moreover, our designed Siamese network, named SigNet, provided better results than  the state-of-the-art results on most of the benchmark signature datasets.
\end{abstract}

%% file: introduction.tex
\section{Introduction}
\label{s:intro}
Signature is one of the most popular and commonly accepted biometric hallmarks that has been used since the ancient times for verifying different entities related to human beings,~\viz~documents, forms, bank checks, individuals, etc. Therefore, signature verification is a critical task and many efforts have been made to remove the uncertainty involved in the manual authentication procedure, which makes \emph{signature verification} an important research line in the field of machine learning and pattern recognition~\cite{Plamondon2000, Impedovo2008}. Depending on the input format, signature verification can be of two types: (1) online and (2) offline. Capturing online signature needs an electronic writing pad together with a stylus, which can mainly record a sequence of coordinates of the electronic pen tip while signing. Apart from the writing coordinates of the signature, these devices are also capable of fetching the writing speed, pressure, etc., as additional information, which are used in the online verification process. On the other hand, the offline signature is usually captured by a scanner or any other type of imaging devices, which basically produces two dimensional signature images. As signature verification has been a popular research topic through decades and substantial efforts are made both on offline as well as on online signature verification purpose.

Online verification systems generally perform better than their offline counter parts~\cite{Munich2003} due to the availability of complementary information such as stroke order, writing speed, pressure,etc. However, this improvement in performances comes at the cost of requiring a special hardware for 
recording the pen-tip trajectory, rising its system cost and reducing the real application scenarios. There are many cases where authenticating offline signature is the only option such as check transaction and document verification. Because of its broader application area, in this paper, we focus on the more challenging task- automatic offline signature verification. Our objective is to propose a convolutional Siamese neural network model to discriminate the genuine signatures and skilled forgeries.

Offline signature verification can be addressed with (1) writer dependent and (2) writer independent approaches~\cite{Bertolini2010}. The writer independent scenario is preferable over writer dependent approaches, as for a functioning system, a writer dependent system needs to be updated (retrained) with every new writer (signer). For a consumer based system, such as bank,  where every day new consumers can open their account this incurs huge cost. Whereas, in writer independent case, a generic system is built to model the discrepancy among the genuine and forged signatures. Training a signature verification system under a writer independent scenario, divides the available signers into train and test sets. For a particular signer, signatures are coupled as \emph{similar} (genuine, genuine) or \emph{dissimilar} (genuine, forged) pairs. From all the tuples of a single signer, equal number of tuples similar and dissimilar pairs are stochastically selected for balancing the number of instances. This procedure is applied to all the signers in the train and test sets to construct the training and test examples for the classifier.

In this regard a signature verifier can be efficiently modelled by a Siamese network which consists of twin convolutional networks accepting two distinct signature images coming from the tuples that are either \emph{similar} or \emph{dissimilar}. The constituting convolutional neural networks (CNN) are then joined by a cost function at the top, which computes a distance metric between the highest level feature representation on each side of the network. The parameters between this twin networks are shared, which in turns guarantees that two extremely similar images could not possibly be mapped by their respective networks to very different locations in feature space because each network computes the same function.

Different hand crafted features have been proposed for offline signature verification tasks. Many of them take into account the global signature image for feature extraction, such as, block codes, wavelet and Fourier series etc~\cite{Kalera2004}. Some other methods consider the geometrical and topological characteristics of local attributes, such as position, tangent direction, blob structure, connected component and curvature~\cite{Munich2003}. Projection and contour based methods~\cite{Dimauro1997} are also quite popular for offline signature verification. Apart from the above mentioned methods, approaches fabricated on direction profile~\cite{Dimauro1997,Ferrer2005}, surroundedness features~\cite{Kumar2012}, grid based methods~\cite{Huang1997a}, methods based on geometrical moments~\cite{Ramesh1999}, and texture based features~\cite{Pal2016} have also become famous in signature verification task. Few structural methods that consider the relations among local features are also explored for the same task. Examples include graph matching~\cite{Chen2006} and recently proposed compact correlated features~\cite{Dutta2016}. On the other hand, Siamese like networks are very popular for different verification tasks, such as, online signature verification~\cite{Bromley1994}, face verification~\cite{Chopra2005,Schroff2015} etc. Furthermore, it has also been used for one-shot image recognition~\cite{Koch2015}, as well as for sketch-based image retrieval task~\cite{Qi2016}. Nevertheless, to the best of our knowledge, till date, convolutional Siamese network has never been used to model an offline signature verifier, which provides our main motivation.

The main contribution of this paper is the proposal of a convolutional Siamese network, named \emph{SigNet}, for offline signature verification problem. This, in contrast to other methods based on hand crafted features, has the ability to model generic signature forgery techniques and many other related properties that envelops minute inconsistency in signatures from the training data. In contrary to other one-shot image verification tasks, the problem with signature is far more complex because of subtle variations in writing styles independent of scripts, which could also encapsulate some degrees of forgery. Here we mine this ultra fine anamorphosis and create a generic model using \emph{SigNet}.

The rest of the paper is organized as follows: In \sect{s:signet} we describe the SigNet and its architechture. \sect{s:expt} presents our experimental validation and compares the proposed method with available state-of-the-art algorithms. Finally, in~\sect{s:concl}, we conclude the paper with a defined future direction.

%% file: signet.tex
\section{SigNet: Siamese Network for Signature Verification}
\label{s:signet}
In this section, at first, the preprocessing performed on signature images is explained in~\sect{ss:preprocessing}. This is followed by a detailed description of the proposed Siamese architecture in~\sect{ss:cnn_siamese}.
%to input to our network.
%We then explain our Siamese network, followed by our network architecture for signature verifications.
%In this section, first we explain the preprocessing of data to input Given a set of signature pairs which are either \emph{similar} or \emph{dissimilar}, we train our model such that the network assigns high level feature vectors closer for \emph{similar} input pairs, and push those away if the input pairs are \emph{dissimilar}.

\subsection{Preprocessing}
\label{ss:preprocessing}
Since batch training a neural network typically needs images of same sizes but the signature images we consider have different sizes ranges from $153 \times 258$ to $819 \times 1137$. We resize all the images to a fixed size $155 \times 220$ using bilinear interpolation. Afterwards, we invert the images so that the background pixels have $0$ values. Furthermore, we normalize each image by dividing the pixel values with the standard deviation of the pixel values of the images in a dataset.

% For all signatures from all the datasets, we apply the same pre-processing strategy. The signatures from the GPDS dataset have a variable size, ranging from $153 \times 258$ pixels to $819 \times 1137$ pixels. Since for training a neural network we need the inputs to have all the same size, we need to normalise the signature images.
% We re-sized the images to a fixed size, using bi-linear interpolation. 
% With this approach, less fine-grained information is lost during the rescaling, specially for the users that have small signatures. On the other hand, the width of the pen strokes becomes inconsistent: for the smaller signatures the pen strokes become much thicker than the pen strokes from the larger signatures.
% Besides resizing the images to a standard size, we also performed the following pre-processing steps:

% \subsubsection*{Inverted the images} We inverted the images so that the white background corresponded to pixel intensity $0$. That is, each pixel of the image is calculated as: $I_inverted(i, j) \leftarrow 255 - I(i, j)$.
% \subsubsection*{Normalized the input} we normalized the input to the neural network by dividing each pixel by the standard deviation of all pixel intensities. We do not normalize the data to have mean 0 another common pre-processing step) since we want the background pixels to be zero-valued.

\subsection{CNN and Siamese Network}
\label{ss:cnn_siamese}
Deep Convolutuional Neural Networks (CNN) are multilayer neural networks consists of several convolutional layers with different kernel sizes interleaved by pooling layers, which summarizes and downsamples the output of its convolutions before feeding to next layers. To get nonlinearity rectified linear units are also used. In this work, we used different convolutional kernels with sizes starting with $11 \times 11$ to $ 3 \times 3$. Generally a differentiable loss function is chosen so that Gradient descent can be applied and the network weights can be optimized. Given a differentiable loss function, the weights of different layers are updated using back propagation. As the optimization can not be applied to all training data where training size is large batch optimizations gives a fair alternative to optimize the network.

%Ideally, a CNN is a multilayer machine learning framework, which consists of an input layer followed by a few convolutional, pooling and fully connected layers. The main aim of a CNN is to model a hierarchy of feature representations and extract high level features for a given pattern. Information in each layer is convolved with a number of filters and further down-sampled by pooling operations, which aggregate values in a small region by functions like max or average operations. The learning of CNN can be done with one of the existing optimisers such as RMSprop, Stochastic Gradient Descent (SGD), etc., which iteratively lowers the defined loss function to train the underlying network in order to produce an accurate feature representation for a given image.

\begin{figure*}
\begin{center}
\includegraphics[width=0.8\textwidth]{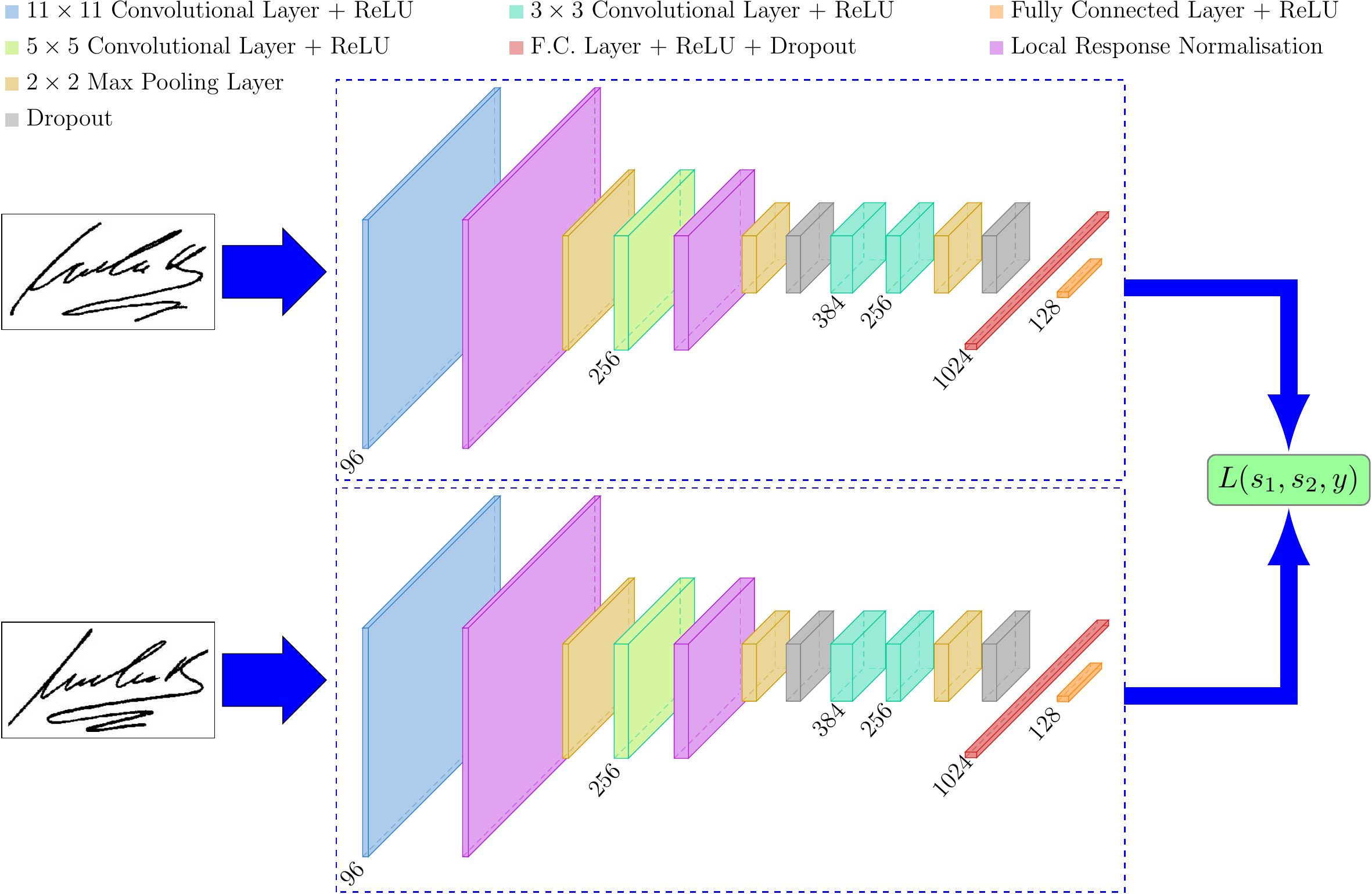}
\end{center}
\caption{Architecture of SigNet: the input layer,~\ie~the $11 \times 11$ convolution layer with ReLU, is shown in blue, whereas all the $3 \times 3$ and $5 \times 5$ convolution layers are depicted in cyan and green respectively. All the local response normalization layers are shown in magenta, all the max pooling layers are depicted in brick color and the dropout layers are exhibited in gray. The last orange block represents the high level feature output from the constituting CNNs, which are joined by the loss function in~\eq{eqn:loss_siam}. (Best viewed in pdf)}
\end{figure*}

Siamese neural network is a class of network architectures that usually contains two identical subnetworks. The twin CNNs have the same configuration with the same parameters and shared weights. The parameter updating is mirrored across both the subnetworks. This framework has been successfully used for dimensionality reduction in weakly supervised metric learning~\cite{Chopra2005} and for face verification in \cite{Schroff2015}. These subnetworks are joined by a loss function at the top, which computes a similarity metric involving the Euclidean distance between the feature representation on each side of the Siamese network. One such loss function that is mostly used in Siamese network is the \emph{contrastive loss}~\cite{Chopra2005} defined as follows:
% The contrastive loss function ($L$) is one of the famous functions that are mostly used of a pair for the pair of features can be formularised as follows:
% Siamese network is designed to contain twin CNNs accepting two different inputs and has been successfully used for dimensionality reduction in weakly supervised metric learning~\cite{Chopra2005}. 

\begin{equation}
\label{eqn:loss_siam}
L(s_1, s_2, y) = \alpha (1-y) D_w^2 + \beta y \max(0, m - D_w)^2
\end{equation}

where $s_1$ and $s_2$ are two samples (here signature images), $y$ is a binary indicator function denoting whether the two samples belong to the same class or not, $\alpha$ and $\beta$ are two constants and $m$ is the margin equal to $1$ in our case. $D_w=\Vert f(s_1;w_1)-f(s_2;w_2) \Vert_2$ is the Euclidean distance computed in the embedded feature space, $f$ is an embedding function that maps a signature image to real vector space through CNN, and $w_1$, $w_2$ are the learned weights for a particular layer of the underlying network. Unlike conventional approaches that assign binary similarity labels to pairs, Siamese network aims to bring the output feature vectors closer for input pairs that are labelled as similar, and push the feature vectors away if the input pairs are dissimilar. Each of the branches of the Siamese network can be seen as a function that embeds the input image into a space. Due to the loss function selected (\eq{eqn:loss_siam}), this space will have the property that images of the same class (genuine signature for a given writer) will be closer to each other than images of different classes (forgeries or signatures of different writers). Both branches are joined together by a layer that computes the Euclidean distance between the two points in the embedded space. Then, in order to decide if two images belong to the similar class (genuine, genuine) or a dissimilar class (genuine, forged) one needs to determine a threshold value on the distance.

%While training a Siamese network with an optimizer, pairs of samples are processed using two identical networks and at the highest level of the network, the loss or error is computed using. Then, the model is back propagated and gradients are computed based on the available sample and the Siamese network weights are updated accordingly.

\subsection{Architecture}
\label{ss:arch}
We have used a CNN architecture that is inspired by Krizhevsky~\etal~\cite{Krizhevsky2012} for an image recognition problem. For the easy reproducibility of our results, we present a full list of parameters used to design the CNN layers in~\tab{tab:cnn_signet}. For convolution and pooling layers, we list the size of the filters as $N \times H \times W$, where $N$ is the number of filters, $H$ is the height and $W$ is the width of the corresponding filter. Here, \emph{stride} signifies the distance between the application of filters for the convolution and pooling operations, and \emph{pad} indicates the width of added borders to the input. Here it is to be mentioned that padding is necessary in order to convolve the filter from the very first pixel in the input image. Throughout the network, we use Rectified Linear Units (ReLU) as the activation function to the output of all the convolutional and fully connected layers. For generalizing the learned features, Local Response Normalization is applied according to~\cite{Krizhevsky2012}, with the parameters shown in the corresponding row in~\tab{tab:cnn_signet}. With the last two pooling layers and the first fully connected layer, we use a Dropout with a rate equal to $0.3$ and $0.5$, respectively.

\begin{table}[ht]
\caption{Overview of the constituting CNNs}
\label{tab:cnn_signet}
\begin{center}
\resizebox{\columnwidth}{!}{
\begin{tabular}{ c c c }
\hline
Layer & Size & Parameters\\
\hline
Convolution & $96 \times 11 \times 11$ & $\text{stride}=1$\\
\multirow{2}{*}{Local Response Norm.} & $\multirow{2}{*}{-}$ & $\alpha=10^{-4}, \beta=0.75$\\
									  &					     & $k=2, n=5$\\
Pooling & $96 \times 3 \times 3$ & $\text{stride}=2$\\
Convolution & $256 \times 5 \times 5$ & $\text{stride}=1, \text{pad}=2$\\
\multirow{2}{*}{Local Response Norm.} & $\multirow{2}{*}{-}$ & $\alpha=10^{-4}, \beta=0.75$\\
									  &					     & $k=2, n=5$\\
Pooling + Dropout & $256 \times 3 \times 3$ & $\text{stride}=2, p=0.3$\\
Convolution & $384 \times 3 \times 3$ & $\text{stride}=1, \text{pad}=1$\\
Convolution & $256 \times 3 \times 3$ & $\text{stride}=1, \text{pad}=1$\\
Pooling + Dropout & $256 \times 3 \times 3$ & $\text{stride}=2, p=0.3$\\
Fully Connected + Dropout & $1024$ & $p=0.5$\\
Fully Connected & $128$ & \\
\hline
\end{tabular}}
\end{center}
\end{table}

\begin{figure*}[thb]
\centering
\resizebox{\textwidth}{!}{
\begin{tabular}{cccccc}
\includegraphics[width=0.16\textwidth]{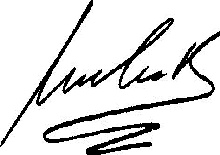} & \includegraphics[width=0.16\textwidth]{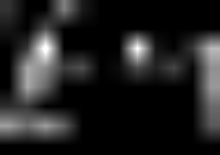} & \includegraphics[width=0.16\textwidth]{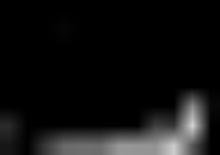} & \includegraphics[width=0.16\textwidth]{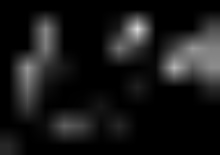} & \includegraphics[width=0.16\textwidth]{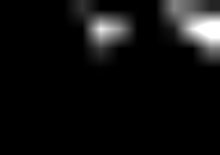} & \includegraphics[width=0.16\textwidth]{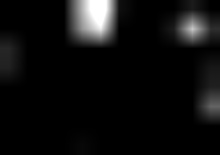}\\
\includegraphics[width=0.16\textwidth]{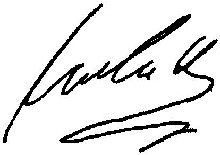} & \includegraphics[width=0.16\textwidth]{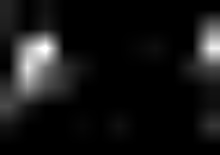} & \includegraphics[width=0.16\textwidth]{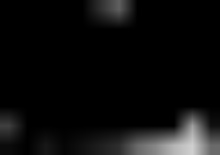} & \includegraphics[width=0.16\textwidth]{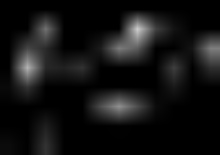} & \includegraphics[width=0.16\textwidth]{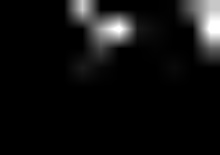} & \includegraphics[width=0.16\textwidth]{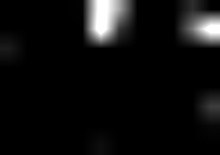}
\end{tabular}}
\caption{A pair of genuine (top left) and forged (bottom left) signatures, and corresponding response maps with five different filters that have produced higher energy activations in the last convolution layer of SigNet.}
\label{fig:activ_filters_layer4}
\end{figure*}

The first convolutional layers filter the $155 \times 220$ input signature image with $96$ kernels of size $11\times 11$ with a stride of $1$ pixels. The second convolutional layer takes as input the (response-normalized and pooled) output of the first convolutional layer and filters it with $256$ kernels of size $5 \times 5$. The third and fourth convolutional layers are connected to one another without any intervention of pooling or normalization of layers. The third layer has $384$ kernels of size $3 \times 3$ connected to the (normalized, pooled, and dropout) output of the second convolutional layer. The fourth convolutional layer has $256$ kernels of size $3 \times 3$. 
This leads to the neural network learning fewer lower level
features  for  smaller  receptive  fields  and  more  features  for
higher level or more abstract features.
The first fully connected layer has $1024$ neurons, whereas the second fully connected layer has $128$ neurons. This indicates that the highest learned feature vector from each side of SigNet has a dimension equal to $128$.

We initialize the weights of the model according to the work of Glorot and Bengio~\cite{Glorot2010}, and the biases equal to $0$. We trained the model using RMSprop for $20$ epochs, using momentum rate equal to $0.9$, and mini batch size equal to $128$. We started with an intial learning rate (LR) equal to $1e-4$ with hyper parameters $\rho=0.9$ and $\epsilon=1e-8$. All these values are shown in~\tab{tab:hyp_param_signet}. Our entire framework is implemented using Keras library with the TensorFlow as backend. The training was done using a GeForce GTX $1070$ and a TITAN X Pascal GPU, and it took $2$ to $9$ hours to run approximately, depending on different databases.

\begin{table}[ht]
\caption{Training Hyper-parameters}
\label{tab:hyp_param_signet}
\begin{center}
\begin{tabular}{ l r }
\hline
Parameter & Value \\
\hline
Initial Learning Rate (LR) & 1e-4\\
Learning Rate Schedule & $\text{LR} \leftarrow \text{LR} \times 0.1$\\
Weight Decay & $0.0005$\\
Momentum $(\rho)$ & $0.9$\\
Fuzz factor $(\epsilon)$ & 1e-8\\
Batch Size & $128$\\
\hline
\end{tabular}
\end{center}
\end{table}

\fig{fig:activ_filters_layer4} shows five different filter activations in the last convolutional layer on a pair of (genuine, forged) signatures, which have received comparatively higher discrimination. The first row corresponds to the genuine signature image, whereas, the second row corresponds to the forged one and these two sigantures are correctly classified as dissimilar by SigNet. Each column starting from the second one shows the activations under the convolution of the same filter. The responses in the respective zones show the areas or signature features that are learned by the network for distinguishing these two signatures.

%% file: experiments.tex
\section{Experiments}
\label{s:expt}
In order to evaluate our signature verification algorithm, we have considered \emph{four} widely used benchmark databases, \viz, (1) CEDAR, (2) GPDS300, (3) GPDS Synthetic Signature Database, and (4) BHSig260 signature corpus. \emph{The source code of SigNet will be available once the paper gets accepted for publication.}

\subsection{Datasets}
\label{sec:datasets}
\subsubsection{CEDAR}
\label{sec:cedar}
CEDAR signature database\footnote{Available at \url{http://www.cedar.buffalo.edu/NIJ/data/signatures.rar}} contains signatures of $55$ signers belonging to various cultural and professional backgrounds. Each of these signers signed $24$ genuine signatures $20$ minutes apart. Each of the forgers tried to emulate the signatures of $3$ persons, $8$ times each, to produce $24$ forged signatures for each of the genuine signers. Hence the dataset comprise $55\times 24=1,320$ genuine signatures as well as $1,320$ forged signatures. The signature images in this dataset are available in gray scale mode.

\subsubsection{GPDS300}
\label{sec:gpds300}
GPDS300 signature corpus\footnote{Available at \url{http://www.gpds.ulpgc.es/download}} comprises $24$ genuine and $30$ forged signatures for $300$ persons. This sums up to $300 \times 24 = 7,200$ genuine signatures and $300 \times 30 = 9,000$ forged signatures. The $24$ genuine signatures of each of the signers were collected in a single day. The genuine signatures are shown to each forger and are chosen randomly from the $24$ genuine ones to be imitated. All the signatures in this database are available in binary form.

\subsubsection{GPDS Synthetic}
\label{sec:gpds_synth}
GPDS synthetic signature database\footnote{Available at \url{http://www.gpds.ulpgc.es/download}} is built based on the synthetic individuals protocol~\cite{Ferrer2013}. This dataset is comprised of $4000$ signers, where each individual has $24$ genuine and $30$ forged signatures resulting in $4000 \times 24 = 96,000$ genuine and $4000 \times 30 = 120,000$ forged signatures.

\subsubsection{BHSig260}
\label{sec:bhsig260}
The BHSig260 signature dataset\footnote{The dataset is available at \url{https://goo.gl/9QfByd}} contains the signatures of $260$ persons, among them $100$ were signed in Bengali and $160$ are signed in Hindi~\cite{Pal2016}. The authors have followed the same protocol as in GPDS300 to generate these signatures. Here also, for each of the signers, $24$ genuine and $30$ forged signatures are available. This results in $100 \times 24 = 2,400$ genuine and $100 \times 30 = 3,000$ forged signatures in Bengali, and $160 \times 24 = 3,840$ genuine and $160 \times 30 = 4,800$ forged signatures in Hindi. Even though this dataset is available together, we experimented with our method separately on the Bengali and Hindi dataset.

\subsection{Performance Evaluation}
A threshold $d$ is used on the distance measure $D(x_i,x_j)$ output by the SigNet to decide whether the signature pair $(i,j)$ belongs to the \emph{similar} or \emph{dissimilar} class. We denote the signature pairs $(i, j)$ with the same identity as $\mathcal{P}_\text{similar}$, whereas all pairs of different identities as $\mathcal{P}_\text{dissimilar}$. Then, we can define the set of all \emph{true positives} (TP) at $d$ as
\begin{equation*}
TP(d) = \lbrace(i, j)\in \mathcal{P}_\text{similar},\text{ with }D(x_i, x_j)\leq d\rbrace
\end{equation*}
Similarly the set of all \emph{true negatives} (TN) at $d$ can be defined as
\begin{equation*}
TN(d) = \lbrace(i, j)\in \mathcal{P}_\text{dissimilar},\text{ with }D(x_i, x_j) > d\rbrace
\end{equation*}
Then the true positive rate $TPR(d)$ and the true negative rate $TNR(d)$ for a given signature, distance $d$ are then defined as
\begin{equation*}
\label{eqn:tpr_tnr}
TPR(d) = \frac{\lvert TP(d)\lvert}{\lvert \mathcal{P}_\text{similar}\lvert},~ TNR(d) = \frac{\lvert TN(d)\lvert}{\lvert \mathcal{P}_\text{dissimilar}\lvert}
\end{equation*}
where $\mathcal{P}_\text{similar}$ is the number of similar signature pairs.
The final accuracy is computed as
\begin{equation}
\text{Accuracy}=\max_{d\in D} \frac{1}{2} (TPR(d) + TNR(d))
\end{equation}
which is the maximum accuracy obtained by varying $d\in D$ from the minimum distance value to the maximum distance value of $D$ with step equal to $0.01$.

\subsection{Experimental Protocol}
 
Since our method is designed for writer independent signature verification, we divide each of the datasets as follows. We randomly select $M$ signers from the $K$ (where $K>M$) available signers of each of the datasets. We keep all the original and forged signatures of these $M$ signers for training and the rest of the $K-M$ signers for testing. Since all the above mentioned datasets contain $24$ genuine signatures for each of the authors, there are only $\nCr{24}{2}=276$ (genuine, genuine) signature pairs available for each author. Similarly, since most of the datasets contain $30$ (for CEDAR $24$) forged signatures for each signer, there are only $24 \times 30 = 720$ (for CEDAR $24 \times 24=576$) (genuine, forged) signature pairs can be obtained for each author. For balancing the similar and dissimilar classes, we randomly choose only $276$ (genuine, forged) signature pairs from each of the writers. This protocol results in $M \times 276$ (genuine, genuine) as well as (genuine, forged) signature pairs for training and $(K-M) \times 276$ for testing. \tab{tab:K_M} shows the values of $K$ and $M$ for different datasets, that are considered for our experiments.

\begin{table}[!htb]
\caption{$K$ and $M$ values of different datasets}
\label{tab:K_M}
\begin{center}
\begin{tabular}{ l c c }
\hline
Datasets & $K$ & $M$ \\
\hline
CEDAR & $55$ & $50$ \\
GPDS300 & $300$ & $150$ \\
GPDS Synthetic & $4000$ & $3200$ \\
Bengali & $100$ & $50$ \\
Hindi & $160$ & $100$ \\
\hline
\end{tabular}
\end{center}
\end{table}

Although most of the existing datasets contain forged signatures, in real life scenarios, there can be cases where getting training samples from forgers might be difficult. Thus, a system trained with genuine-forged signature pairs will be inadequate to deal with such set up. One way to deal with this type of situations is to use only the genuine signatures of other signer as forged signatures (called as \emph{unskilled} forged signatures). To be applicable in such scenarios, we have performed an experiment only on the GPDS-300 dataset, where the genuine signatures of other writers are used as unskilled forged signatures. However, during testing, we have used genuine-forged pairs of the same signers, \ie, we tested our system for it's ability to distinguish between genuine and forged signatures of the same person. % In~\tab{tab:perf-comp}, we provide accuracy for both the training modalities namely genuine-genuine and genuine-forged for GPDS-300 dataset.

\begin{table*}[!thb]
\caption{Comparison of the proposed method with the state-of-the-art methods on various signature databases.}
\label{tab:perf-comp}
\begin{center}
\resizebox{0.77\textwidth}{!}{
\begin{tabular}{|c c c c c c|}
\hline
Databases & State-of-the-art Methods & \#Signers & Accuracy & FAR & FRR\\
\hline
\multirow{6}{*}{CEDAR Signature Database} & Word Shape (GSC) (Kalera~\etal~\cite{Kalera2004}) & $55$ & $78.50$ & $19.50$ & $22.45$\\
& Zernike moments (Chen and Srihari~\cite{Chen2005}) & $55$ & $83.60$ & $16.30$ & $16.60$\\
& Graph matching (Chen and Srihari~\cite{Chen2006}) & $55$ & $92.10$ & $8.20$ & $7.70$\\
& Surroundedness features (Kumar~\etal~\cite{Kumar2012}) & $55$ & $91.67$ & $8.33$ & $8.33$\\
& Dutta~\etal~\cite{Dutta2016} & $55$ & $\mathbf{100.00}$ & $\mathbf{0.00}$ & $\mathbf{0.00}$\\
& SigNet & $55$ & $\mathbf{100.00}$ & $\mathbf{0.00}$ & $\mathbf{0.00}$\\
\hline
\multirow{6}{*}{GPDS 300 Signature Corpus} & Ferrer~\etal~\cite{Ferrer2005} & $160$ & $86.65$ & $12.60$ & $14.10$\\
& Vargas~\etal~\cite{Vargas2007} & $160$ & $87.67$ & $14.66$ & $10.01$\\
& Solar~\etal~\cite{Solar2008} & $160$ & $84.70$ & $14.20$ & $16.40$\\
& Kumar~\etal~\cite{Kumar2012} & $300$ & $86.24$ & $13.76$ & $13.76$\\
& Dutta~\etal~\cite{Dutta2016} & $300$ & $\mathbf{88.79}$ & $\mathbf{11.21}$ & $\mathbf{11.21}$\\
& SigNet & $300$ & $76.83$ & $23.17$ & $23.17$\\
& SigNet (unskilled forged) & $300$ & $65.36$ & $34.64$ & $34.64$\\
\hline
\multirow{2}{*}{GPDS Synthetic Signature Corpus} & Dutta~\etal~\cite{Dutta2016} & $4000$ & $73.67$ & $28.34$ & $27.62$\\
& SigNet & $4000$ & $\mathbf{77.76}$ & $\mathbf{22.24}$ & $\mathbf{22.24}$\\
\hline
\multirow{3}{*}{Bengali} & Pal~\etal~\cite{Pal2016} & $100$ & $66.18$ & $33.82$ & $33.82$\\
& Dutta~\etal~\cite{Dutta2016} & $100$ & $84.90$ & $15.78$ & $14.43$ \\
& SigNet & $100$ & $\mathbf{86.11}$ & $\mathbf{13.89}$ & $\mathbf{13.89}$\\
\hline
\multirow{3}{*}{Hindi} & Pal~\etal~\cite{Pal2016} & $100$ & $75.53$ & $24.47$ & $24.47$\\
& Dutta~\etal~\cite{Dutta2016} & $100$ & $\mathbf{85.90}$ & $\mathbf{13.10}$ & $\mathbf{15.09}$ \\
& SigNet & $100$ & $84.64$ & $15.36$ & $15.36$\\
\hline
\end{tabular}}
\end{center}
\end{table*}
\subsection{Results and Discussions}
\label{sec:disc}
\tab{tab:perf-comp} shows the accuracies of our proposed SigNet together with other state-of-the-art methods on different datasets discussed in~\sect{sec:datasets}. It is to be noted that SigNet outperformed the state-of-the-art methods on three datasets,~\viz~GPDS Synthetic, Bengali, and CEDAR dataset. A possible reason for the lower performance on the GPDS300 is the less number of signature samples for learning with many different signature styles. However, on GPDS Synthetic, our proposed network outperformed the same method proposed by Dutta~\etal~\cite{Dutta2016} possibly because there were plenty of training samples for learning the available signature styles. Moreover, it can be observed that our system trained on genuine-unskilled forged pairs is outperformed by the system trained on genuine-forged examples. This is quite justified and very intuitive as identifying forgeries of a signature needs attention to minute details of one's signature, which can not be captured when unskilled forged signatures (\ie genuine signatures of other signers) are used as training examples.

\begin{figure}[htb]
\centering
\includegraphics[width=\columnwidth]{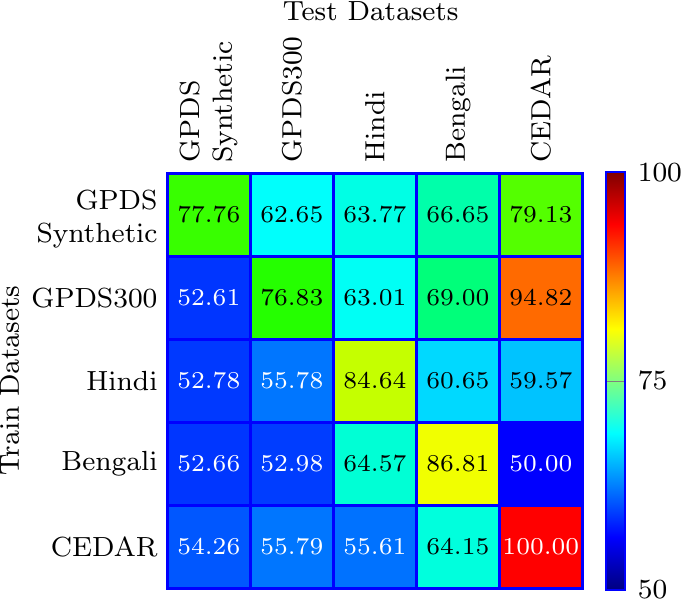}
\caption{Accuracies obtained by \emph{SigNet} with cross dataset settings.}
\label{fig:acc_cross_data}
\end{figure}

To get some ideas on the generalization of the proposed network and the strength of the models learned on different datasets, we performed a second experiment with cross dataset settings. To do this, at a time, we have trained a model on one of the above mentioned datasets and tested it on all the other corpus. We have repeated this same process over all the datasets. The accuracies obtained by SigNet on the cross dataset settings are shown in~\fig{fig:acc_cross_data}, where the datasets used for training are indicated in rows and the datasets used for testing are exhibited along columns. It is to be observed that for all the datasets, the highest accuracy is obtained with a model trained on the same dataset. This implies all the datasets have some distinctive features, despite the fact that, CEDAR, GPDS300 and GPDS Synthetic datasets contain signatures with nearly same style (some handwritten letters with initials etc.). However, this fact is justifiable in case of BHSig260 dataset, because it contains signatures in Indic script and the signatures generally look like normal text containing full names of persons. Therefore, it is probable that the network models some script  based features in this case. Furthermore, it is usually noted that the system trained on a comparatively bigger and diverse dataset is more robust than the others, which is the reason why better average accuracies are obtained by the model trained on GPDS Synthetic and GPDS300. These experiments strongly show the possibility of creating signature verification system in those cases where training is not possible due to the dearth of sufficient data. In those situation, a robust pretrained model could be used with a lightweight fine tuning on the available specific data.
We also thought of the situation where the forger not knowing the real identity of the person, he or she introduces his signature or scribbling which has more variations than the skilled forged ones. To evaluate this, we used the trained model on GPDS300 (trained with skilled forgery) and tested it on signatures placed against a random forgery (\ie genuine signature of another person) giving an expected increase in performance with $79.19\%$ accuracy rate in GPDS300 dataset (keeping rest of the experimental setup same). This also proves that the model trained to find subtle differences in signature, also performs well when the variations in signatures are large.
% Hindi = $84.64\%$
% Bengali = $86.81\%$
% GPDS300 = $76.83\%$
% GPDS960 = $77.76\%$
% CEDAR = $100\%$

% \begin{tabular}{cc|c|c|c|c|c|l}
% \cline{3-7}
% & & \multicolumn{5}{ c| }{Test} \\ \cline{3-7}
% & & GPDS960 & GPDS300 & HINDI & BENGALI & CEDAR \\ \cline{1-7}
% \multicolumn{1}{ |c  }{\multirow{5}{*}{Train} } &
% \multicolumn{1}{ |c| }{GPDS960} & \textbf{77.76} & 62.65 & 63.77 & 66.65 & 79.13    \\ \cline{2-7}
% \multicolumn{1}{ |c  }{}                        &
% \multicolumn{1}{ |c| }{GPDS300} & 52.61 & \textbf{76.83} & 63.01 & 69 & 94.82    \\ \cline{2-7}
% \multicolumn{1}{ |c  }{}                        &
% \multicolumn{1}{ |c| }{HINDI} & 52.78 & 55.78 & \textbf{84.64} & 60.65 & 59.57    \\ \cline{2-7}
% \multicolumn{1}{ |c  }{}                        &
% \multicolumn{1}{ |c| }{BENGALI} & 52.66 & 52.98 & 64.57 & \textbf{86.81} & 50    \\ \cline{2-7}
% \multicolumn{1}{ |c  }{}                        &
% \multicolumn{1}{ |c| }{CEDAR} & 54.26 & 55.79 & 55.61 & 64.15 & \textbf{100}    \\ \cline{1-7}
% \end{tabular}

%% file: conclusions.tex
\section{Conclusions}
\label{s:concl}
In this paper, we have presented a framework based on Siamese network for offline signature verification, which uses writer independent feature learning. This method does not rely on hand-crafted features unlike its predecessors, instead it learns them from data in a writer independent scenario. Experiments conducted on GPDS Syntehtic dataset demonstrate that this is a step towards modelling a generic prototype for real forgeries based on synthetically generated data. Also, our experiments made on cross domain datasets emphasize how well our architecture models the fraudulence of different handwriting style of different signers and forgers with diverse background and scripts. Furthermore, the SigNet designed by us has surpassed the state-of-the-art results on most of the benchmark Signature datasets, which is encouraging for further research in this direction. Our future work in this line will focus on the development of more enriched network model. Furthermore, other different frameworks for verification task will also be explored.

% These are very interesting findings and it works so well. Future work can explore how far this idea can be extended.

% In this paper, we have proposed SigNet: a Siamese network consisting twin convolutional networks with several convolution, max pooling and dropout layers.

% SigNet accepts a tuple of two different signature images from the tuple of images that are wither similar (original, original) or dissimilar (original, forged). These two constituting networks are joined by an energy function at the top, which computes a distance metric between the highest level feature representation on each side of the Siamese network. 
% Furthermore, in each iteration of the optimization process, pairs of images are processed, loss is computed and minimized, which trains the network to assign high level output feature vectors closer for similar input pairs, and push those away if the input pairs are dissimilar. Our experiments made on cross domain datasets emphasize how well our architecture models the fraudulence of different handwriting style of different signers and forgers with diverse background. Furthermore, the SigNet designed by us has surpassed the state-of-the-art results on most of the benchmark Signature datasets, which is encouraging for further research in this direction.

%% file: acknowledgement.tex
\section*{Acknowledgement}
\label{sec:ack}
This work has been partially supported by the European Union's research and innovation program under the Marie Sk\l{}odowska-Curie grant agreement No. 665919. The TITAN X Pascal GPU used for this research was donated by the NVIDIA Corporation.

%% file: PRLetters_SigNet.bbl
\begin{thebibliography}{10}
\expandafter\ifx\csname url\endcsname\relax
  \def\url#1{\texttt{#1}}\fi
\expandafter\ifx\csname urlprefix\endcsname\relax\def\urlprefix{URL }\fi
\expandafter\ifx\csname href\endcsname\relax
  \def\href#1#2{#2} \def\path#1{#1}\fi

\bibitem{Plamondon2000}
R.~Plamondon, S.~Srihari, Online and off-line handwriting recognition: a
  comprehensive survey, IEEE TPAMI 22~(1) (2000) 63--84.

\bibitem{Impedovo2008}
D.~Impedovo, G.~Pirlo, Automatic signature verification: The state of the art,
  IEEE TSMC 38~(5) (2008) 609--635.

\bibitem{Munich2003}
M.~E. Munich, P.~Perona, Visual identification by signature tracking, IEEE
  TPAMI 25~(2) (2003) 200--217.

\bibitem{Bertolini2010}
D.~Bertolini, L.~Oliveira, E.~Justino, R.~Sabourin, Reducing forgeries in
  writer-independent off-line signature verification through ensemble of
  classifiers, PR 43~(1) (2010) 387--396.

\bibitem{Kalera2004}
M.~K. Kalera, S.~N. Srihari, A.~Xu, Offline signature verification and
  identification using distance statistics, IJPRAI 18~(7) (2004) 1339--1360.

\bibitem{Dimauro1997}
G.~Dimauro, S.~Impedovo, G.~Pirlo, A.~Salzo, A multi-expert signature
  verification system for bankcheck processing, IJPRAI 11~(05) (1997) 827--844.

\bibitem{Ferrer2005}
M.~A. Ferrer, J.~B. Alonso, C.~M. Travieso, Offline geometric parameters for
  automatic signature verification using fixed-point arithmetic, IEEE TPAMI
  27~(6) (2005) 993--997.

\bibitem{Kumar2012}
R.~Kumar, J.~Sharma, B.~Chanda, Writer-independent off-line signature
  verification using surroundedness feature, PRL 33~(3) (2012) 301--308.

\bibitem{Huang1997a}
K.~Huang, H.~Yan, Off-line signature verification based on geometric feature
  extraction and neural network classification, PR 30~(1) (1997) 9--17.

\bibitem{Ramesh1999}
V.~Ramesh, M.~N. Murty, Off-line signature verification using genetically
  optimized weighted features, PR 32~(2) (1999) 217--233.

\bibitem{Pal2016}
S.~Pal, A.~Alaei, U.~Pal, M.~Blumenstein, Performance of an off-line signature
  verification method based on texture features on a large indic-script
  signature dataset, in: DAS, 2016, pp. 72--77.

\bibitem{Chen2006}
S.~Chen, S.~Srihari, A new off-line signature verification method based on
  graph, in: ICPR, 2006, pp. 869--872.

\bibitem{Dutta2016}
A.~Dutta, U.~Pal, J.~Llad\'{o}s, Compact correlated features for writer
  independent signature verification, in: ICPR, 2016, pp. 3411--3416.

\bibitem{Bromley1994}
J.~Bromley, I.~Guyon, Y.~LeCun, E.~S\"{a}ckinger, R.~Shah, Signature
  verification using a "siamese" time delay neural network, in: NIPS, 1994, pp.
  737--744.

\bibitem{Chopra2005}
S.~Chopra, R.~Hadsell, Y.~LeCun, Learning a similarity metric discriminatively,
  with application to face verification, in: CVPR, 2005, pp. 539--546.

\bibitem{Schroff2015}
F.~Schroff, D.~Kalenichenko, J.~Philbin, Facenet: A unified embedding for face
  recognition and clustering, in: CVPR, 2015, pp. 815--823.

\bibitem{Koch2015}
G.~Koch, R.~Zemel, R.~Salakhutdinov, Siamese neural networks for one-shot image
  recognition, in: ICML, 2015, pp. 1--8.

\bibitem{Qi2016}
Y.~Qi, Y.~Z. Song, H.~Zhang, J.~Liu, Sketch-based image retrieval via siamese
  convolutional neural network, in: ICIP, 2016, pp. 2460--2464.

\bibitem{Krizhevsky2012}
A.~Krizhevsky, I.~Sutskever, G.~E. Hinton, Imagenet classification with deep
  convolutional neural networks, in: NIPS, 2012, pp. 1097--1105.

\bibitem{Glorot2010}
X.~Glorot, Y.~Bengio, Understanding the difficulty of training deep feedforward
  neural networks, in: AISTATS, 2010, pp. 249--256.

\bibitem{Ferrer2013}
M.~A. Ferrer, M.~Diaz-Cabrera, A.~Morales, Synthetic off-line signature image
  generation, in: ICB, 2013, pp. 1--7.

\bibitem{Chen2005}
S.~Chen, S.~Srihari, Use of exterior contours and shape features in off-line
  signature verification, in: ICDAR, 2005, pp. 1280--1284.

\bibitem{Vargas2007}
F.~Vargas, M.~Ferrer, C.~Travieso, J.~Alonso, Off-line handwritten signature
  gpds-960 corpus, in: ICDAR, Vol.~2, 2007, pp. 764--768.

\bibitem{Solar2008}
J.~Ruiz~del Solar, C.~Devia, P.~Loncomilla, F.~Concha, Offline signature
  verification using local interest points and descriptors, in: CIARP, 2008,
  pp. 22--29.

\end{thebibliography}
